\documentclass[final]{cvpr}

\usepackage{times}
\usepackage{epsfig}
\usepackage{graphicx}
\usepackage{amsmath}
\usepackage{amssymb}
\usepackage{pifont}% http://ctan.org/pkg/pifont
\usepackage{booktabs}
\usepackage{multirow}

\usepackage[pagebackref=true, breaklinks=true, letterpaper=true, colorlinks,citecolor=citecolor, linkcolor=linkcolor, bookmarks=false]{hyperref}

\usepackage[table,xcdraw]{xcolor}
\usepackage{enumitem}
\usepackage{wrapfig}

\def\cvprPaperID{7261} % *** Enter the CVPR Paper ID here
\def\confYear{CVPR 2022}
\newcommand{\cmark}{\ding{51}}%
\newcommand{\authorskip}{\hspace{5mm}}
\definecolor{green}{rgb}{0.0,0.8,0.02}
\definecolor{citecolor}{HTML}{0071BC}
\definecolor{linkcolor}{HTML}{ED1C24}
\newlength\savewidth
\begin{document}

%%%%%%%%% TITLE
\title{Learning Affinity from Attention: End-to-End Weakly-Supervised Semantic Segmentation with Transformers}

%\title{Affinity from Attention: End-to-End Weakly-Supervised Semantic Segmentation with Transformers}

\author{Lixiang Ru$^{1}$ \authorskip Yibing Zhan$^{2}$ \authorskip Baosheng Yu$^{3}$ \authorskip Bo Du$^{1}$\thanks{Corresponding author.}
\\
$^{1}$ School of Computer Science, Wuhan University, China\\
$^{2}$ JD Explore Academy, China \authorskip\authorskip
$^{3}$ The University of Sydney, Australia
\\
{\tt\small {\{rulixiang, dubo\}@whu.edu.cn}\authorskip zhanyibing@jd.com\authorskip baosheng.yu@sydney.edu.au}
}

\maketitle
%%%%%%%%% ABSTRACT
\begin{abstract}
    Weakly-supervised semantic segmentation (WSSS) with image-level labels is an important and challenging task. Due to the high training efficiency, end-to-end solutions for WSSS have received increasing attention from the community. However, current methods are mainly based on convolutional neural networks and fail to explore the global information properly, thus usually resulting in incomplete object regions. In this paper, to address the aforementioned problem, we introduce Transformers, which naturally integrate global information, to generate more integral initial pseudo labels for end-to-end WSSS. Motivated by the inherent consistency between the self-attention in Transformers and the semantic affinity, we propose an Affinity from Attention (AFA) module to learn semantic affinity from the multi-head self-attention (MHSA) in Transformers. The learned affinity is then leveraged to refine the initial pseudo labels for segmentation. In addition, to efficiently derive reliable affinity labels for supervising AFA and ensure the local consistency of pseudo labels, we devise a Pixel-Adaptive Refinement module that incorporates low-level image appearance information to refine the pseudo labels. We perform extensive experiments and our method achieves 66.0\% and 38.9\% mIoU on the PASCAL VOC 2012 and MS COCO 2014 datasets, respectively, significantly outperforming recent end-to-end methods and several multi-stage competitors. Code is available at \url{https://github.com/rulixiang/afa}.

\end{abstract}

%%%%%%%%% BODY TEXT
\section{Introduction}

\begin{figure}[!tp]
    \centering
    \includegraphics[width=0.45\textwidth]{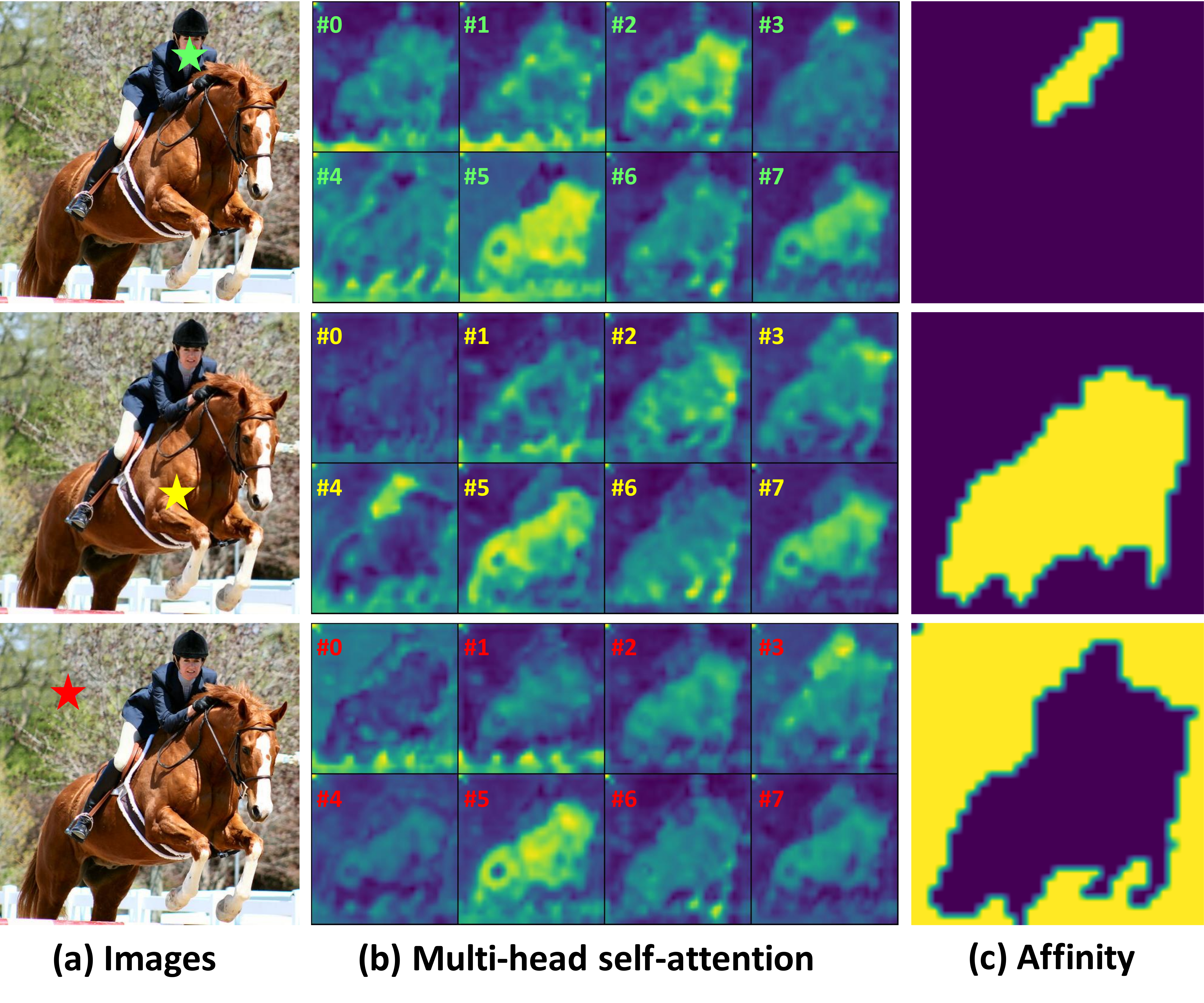}
    \caption{(a) Image and the query points (denote with "$\bigstar$") to visualize the attention and affinity maps; (b) the self-attention maps in Transformer blocks only capture coarse semantic-level affinity relations; (c) the learned reliable semantic affinity from self-attention with our proposed method.}
    \vspace{-4mm}
    \label{fig_intro}
\end{figure}

\par Semantic segmentation, aiming at labeling each pixel in an image, is a fundamental task in vision. In the past decade, deep neural networks have achieved great success in semantic segmentation. However, due to the data-hungry nature of deep neural networks, fully-supervised semantic segmentation models usually require a large amount of data with labour intensive pixel-level annotations. To settle this problem, some recent methods seek to devise semantic segmentation models using weak/cheap labels, such as image-level labels \cite{ahn2018learning,lee2021railroad,wu2021embedded,lee2021anti,xu2021leveraging,li2021pseudo,ru2021learning}, points \cite{akiva2021towards}, scribbles \cite{lin2016scribblesup,zhang2021dynamic,zhang2021affinity}, and bounding boxes \cite{lee2021bbam}. Our method falls into the category of weakly-supervised semantic segmentation (WSSS) using only image-level labels, which is the most challenging one in all WSSS scenarios.

\par Prevailing WSSS methods with image-level labels commonly adopt a multi-stage framework \cite{ru2021learning,lee2021anti,lee2021reducing}. Specifically, these methods firstly train a classification model and then generate Class Activation Maps (CAM) \cite{zhou2016learning} as the pseudo labels. After refinement, the pseudo labels are leveraged to train a standalone semantic segmentation network as the final model. This multi-stage framework needs to train multiple models for different purposes, thus obviously complicating the training streamline and slowing down the efficiency. To avoid this problem, several end-to-end solutions have been recently proposed for WSSS \cite{araslanov2020single,zhang2021affinity,zhang2020reliability,akiva2021towards}. However, these methods are commonly based on convolution neural networks and fail to explore the global feature relations properly, which turns out to be crucial for activating integral object regions~\cite{gao2021ts}, thus significantly affecting the quality of generated pseudo labels.

\par
\begin{figure}[!tp]
    \centering
    \includegraphics[width=0.45\textwidth]{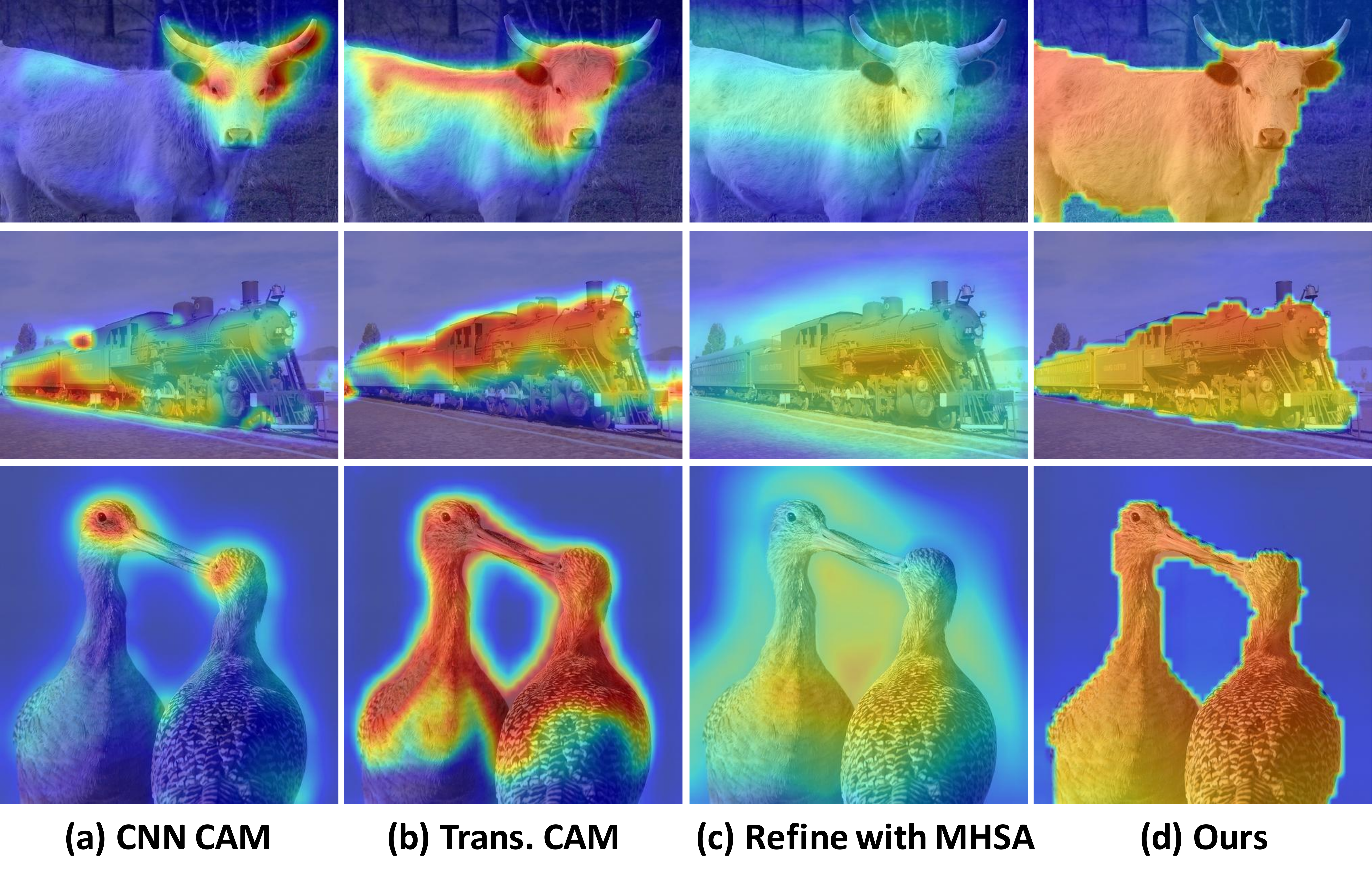}
    \caption{CAM generated with (b) Transformers activates more integral regions than (a) CNN. Refining CAM with (c) coarse MHSA doesn't work well, while (d) the learned affinity could remarkably improve the generated CAM.}
    \vspace{-4mm}
    \label{fig_intro_cam}
\end{figure}

 Recently, Transformers \cite{vaswani2017attention} have achieved significant breakthroughs in numerous visual applications \cite{xie2021segformer,zheng2021rethinking,arnab2021vivit}. We argue that the Transformer architecture naturally benefits the WSSS task. Firstly, the self-attention mechanism in Transformers could model the global feature relations and conquer the aforementioned drawback of convolutional neural networks, thus discovering more integral object regions. As shown in Fig.~\ref{fig_intro}, we find that the multi-head self-attention (MHSA) in Transformers could capture semantic-level affinity, and can thus be used to improve the coarse pseudo labels. However, the affinity captured in MHSA is still inaccurate (Fig.~\ref{fig_intro}(b)), \ie, directly applying MHSA as affinity to revise the labels does not work well in practice, which is shown in Fig.~\ref{fig_intro_cam} (c). %% need examples????

\par  Based on the above analysis, we propose a Transformer-based end-to-end framework for WSSS. Specifically, we leverage Transformers to generate CAM as the initial pseudo labels, to avoid the intrinsic drawback of convolutional neural networks. We further exploit the inherent affinity in Transformer blocks to improve the initial pseudo labels. Since the semantic affinity in MHSA is coarse, we propose an Affinity from Attention (AFA) module, which aims to derive reliable pseudo affinity labels to supervise the semantic affinity learned from the MHSA in Transformer. The learned affinity is then employed to revise the initial pseudo labels via random walk propagation \cite{ahn2018learning,ahn2019weakly}, which could diffuse object regions and dampen the falsely activated regions. To derive highly-confident pseudo affinity labels for AFA and ensure the local consistency of the propagated pseudo labels, we further propose a Pixel-Adaptive Refinement module (PAR). Based on the pixel-adaptive convolution \cite{araslanov2020single,su2019pixel}, PAR efficiently integrates the RGB and position information of local pixels to refine the pseudo labels, enforcing better alignment with low-level image appearance. In addition, given the simplicity, our model can be trained in an end-to-end manner, thus avoiding a complex training pipeline. Experimental results on PASCAL VOC 2012 \cite{everingham2010pascal} and MS COCO 2014 \cite{lin2014microsoft} demonstrate that our method remarkably surpasses recent end-to-end methods and several multi-stage competitors.

\par In summary, our contributions are listed as follows.
\begin{itemize}[noitemsep,nolistsep,leftmargin=*]
    \item We propose an end-to-end Transformer-based framework for WSSS with image-level labels. To the best of our knowledge, this is the first work to explore Transformers for WSSS.
    \item We exploit the inherent virtue of Transformer and devise an Affinity from Attention (AFA) module. AFA learns reliable semantic affinity from MHSA and propagates the pseudo labels with the learned affinity.
    \item We propose an efficient Pixel-Adaptive Refinement (PAR) module, which incorporates the RGB and position information of local pixels for label refinement.
\end{itemize}

\section{Related Work}

\subsection{Weakly-Supervised Semantic Segmentation}
\noindent\textbf{Multi-stage Methods.} Most WSSS methods with image-level labels are accomplished in a multi-stage process. Commonly, these approaches train a classification network to produce the initial pseudo pixel-level labels with CAM. To address the drawback of incomplete object activation of CAM, \cite{wei2017object,zhang2021complementary,sun2021ecs} utilize \texttt{Erasing} strategy to erase the most discriminative regions and thus discover more complete object regions. Inspired by the observation that the classification network tends to focus on different object regions at different training stages, \cite{jiang2019integral,yao2021non,kim2021discriminative} accumulate the activated regions in the training process. \cite{li2021group,sun2020mining,wu2021embedded} propose to mine semantic regions from multiple input images, discovering similar semantic regions. A prevailing group of WSSS trains classification network with auxiliary tasks to ensure integral object discovery \cite{wang2020self,chang2020weakly,ru2021learning,ru2022weakly}. Some recent researches interpret CAM generation from novel perspectives, such as causal inference \cite{zhang2020causal}, information bottleneck theory \cite{lee2021reducing}, and anti-adversarial attack \cite{lee2021anti}.

\begin{figure*}[htp]
    \centering
    \includegraphics[width=0.95\textwidth]{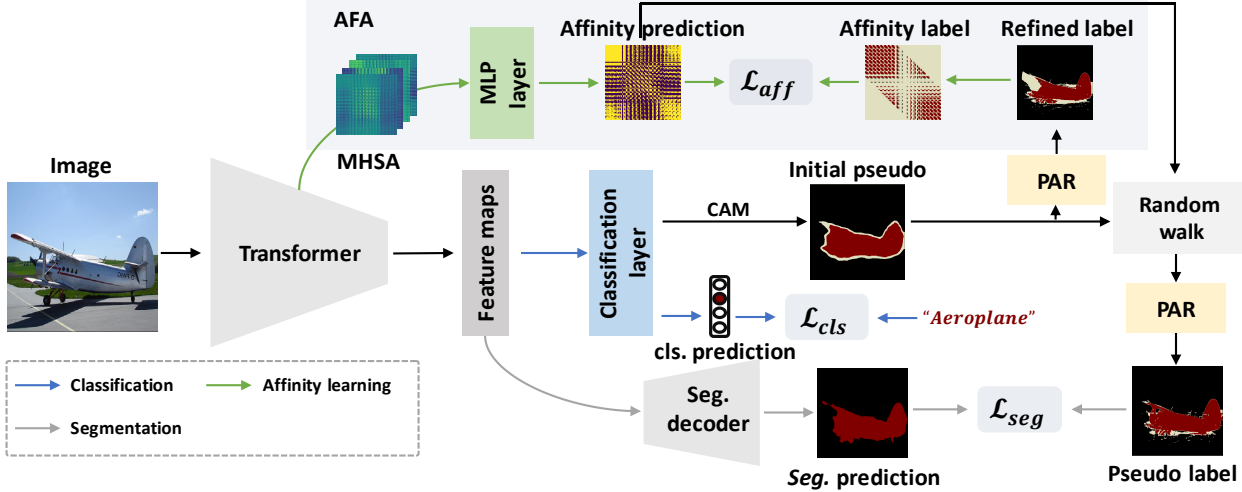}
    \vspace{-2mm}
    \caption{The proposed end-to-end framework for WSSS. We use a Transformer backbone as the encoder to extract feature maps. The initial pseudo labels are generated with CAM \cite{zhou2016learning} and then refined with the proposed PAR. In the AFA module, we derive the semantic affinity from MHSA in Transformer blocks. AFA is supervised with the pseudo affinity labels derived from the refined label. Next, we employ the learned affinity to revise the pseudo labels via random walk propagation \cite{ahn2018learning,ahn2019weakly}. The propagated labels are finally refined with PAR as the pseudo labels for the segmentation branch. }
    \vspace{-2mm}
    \label{fig_wetr}
\end{figure*}

\noindent\textbf{End-to-End Methods.} Due to the extremely limited supervision, training an end-to-end model for WSSS with favorable performance is difficult. \cite{papandreou2015weakly} proposes an adaptive Expectation-Maximization framework to infer pseudo ground truth for segmentation. \cite{pinheiro2015image} tackles WSSS with image-level labels as a multi-instance learning (MIL) problem and devises the \texttt{Log-Sum-Exp} aggregation function to drive the network to assign correct pixel labels.
Incorporating \texttt{nGWP} pooling, pixel-adaptive mask refinement, and stochastic low-level information transfer, 1Stage \cite{araslanov2020single} achieves comparable performance with multi-stage models. In \cite{zhang2020reliability}, RRM takes CAM as the initial pseudo labels and employs CRF \cite{krahenbuhl2011efficient} to produce the refined label as supervision for segmentation. RRM also introduces an auxiliary regularization loss \cite{tang2018regularized} to ensure the consistency between segmentation map and low-level image appearance. \cite{zhang2021adaptive} introduces the adaptive affinity field \cite{ke2018adaptive} with weighted affinity kernels and {feature-to-prototype} alignment loss to ensure the semantic fidelity. The above methods commonly adopt CNN and raise the inherent drawback of convolution, \ie, failing to capture the global information, leading to the incomplete activation of objects \cite{gao2021ts}. In this work, we explore Transformers for end-to-end WSSS to address this issue.

\subsection{Transformer in Vision}
\par In \cite{dosovitskiy2020image}, Dosovitskiy \etal proposes Vision Transformer (ViT), the first work to apply pure Transformer architecture for visual recognition tasks, achieving astonishing performance on visual classification benchmarks. Later variants show ViT also benefits downstream vision tasks, such as semantic segmentation \cite{xie2021segformer,cheng2021per,zheng2021rethinking}, depth estimation \cite{ranftl2021vision}, and video understanding \cite{arnab2021vivit}. In \cite{gao2021ts}, Gao \etal proposes the first Transformers-based method (TS-CAM) for weakly-supervised object localization (WSOL). Close to WSSS, WSOL aims at localizing objects with only image-level supervision. TS-CAM trains a ViT model with image-level supervision, generates semantic-aware CAM, and couples the generated CAM with semantic-agnostic attention maps. The semantic-agnostic attention maps are derived from the attention of \texttt{class token} against other patch tokens. Nevertheless, TS-CAM didn't exploit the intrinsic semantic affinity in MHSA to promote the localization results. In this work, we propose to learn the reliable semantic affinity from MHSA and propagate CAM with the learned affinity. %% 

\section{Methodology}
\par In this section, we first introduce the Transformer backbone and CAM to generate initial pseudo labels. We then present the Affinity from Attention (AFA) module to learn reliable semantic affinity and propagate the initial pseudo labels with the learned affinity. Afterward, we introduce a Pixel-Adaptive Refinement (PAR) module to ensure the local consistency of pseudo labels. The overall loss function for optimization is presented in Section~\ref{sec_loss}.

\subsection{Transformer Backbone}
\par As shown in Fig.~\ref{fig_wetr}, our framework uses the Transformers as the backbone. An input image is firstly split into $h\times w$ patches, where each patch is flattened and linearly projected to form $h\times w$ tokens. In each Transformer block, the multi-head self-attention (MHSA) is used to capture global feature dependencies. Specifically, for the $i^{th}$ head, patch tokens are projected with Multi-Layer Perception (MLP) layers and construct queries $Q_i \in \mathbb{R}^{hw\times d_{k}}$, keys $K_i \in \mathbb{R}^{hw\times d_{k}}$, and values $V_i \in \mathbb{R}^{hw\times d_{v}}$. $d_{k}$ is feature dimension of queries and keys, and $d_{v}$ denotes the feature dimension of values. Based on $Q_i$, $K_i$ and $V_i$, the self-attention matrix ${S_i}$ and outputs $X_i$ are
\begin{equation}
    \label{eq_mhsa}
    S_i = \frac{{Q_i}{K_i}^\top}{\sqrt{d_k}}, \quad X_{i} = \mathtt{softmax}({S_i})V_i.
\end{equation}
The final output $X_o$ of the Transformer block is constructed by feeding $(X_1\|X_2\|...\|X_n)$ into feed-forward layers (FFN), \ie, $X_o = \mathtt{FFN}(X_1\|X_2\|...\|X_n)$, where $\mathtt{FFN}(\cdot)$ consists of Layer Normalization \cite{ba2016layer} and MLP layers. $(\cdot\|\cdot)$ denotes the concatenation operation. By stacking multiple Transformer blocks, the backbone produces feature maps for the subsequent modules.

\subsection{CAM Generation}
\par Considering the simplicity and inference efficiency, we adopt class activation maps (CAM) \cite{zhou2016learning} as the initial pseudo labels. For the extracted feature maps $F\in \mathbb{R}^{hw\times d}$ and a given class $c$, the activation map $M^c$ is generated via weighting the feature maps in $F$ with their contribution to class $c$, \ie, the weight matrix $W$ in the classification layer,
\begin{equation}
    \label{eq_cam}
    \small
    \begin{split}
        {M}^c&=\mathtt{ReLu}\Bigl(\sum_{i=1}^d {W}^{i,c}{F}^{i}\Bigr),
    \end{split}
\end{equation}
where \texttt{ReLu} function is used to remove the negative activations. \texttt{Min-Max} normalization is applied to scale $M^c$ to $[0,1]$. A background score $\beta$ ($0<\beta<1$) is then used to differentiate foreground and background regions.

\subsection{Affinity from Attention} \label{section_afa}
\par As shown in Fig.~\ref{fig_intro}, we notice the consistency between MHSA in Transformers and the semantic-level affinity, which motivates us to use MHSA to discover the object regions. However, since no explicit constraints are imposed on self-attention matrices during the training process, the learned affinity in MHSA is typically coarse and inaccurate, which means directly applying MHSA as affinity to refine the initial labels does not work well (Fig~\ref{fig_intro_cam} (c)). Therefore, we propose the Affinity from Attention module (AFA) to counter this problem.% and accomplish label refinement with affinity.

\par Assuming MHSA in a Transformer block is denoted as $S\in\mathbb{R}^{hw\times hw\times n}$, where $hw$ is the flattened spatial size and $n$ is the number of attention heads. In our AFA module, we directly produce the semantic affinity by linearly combining the multi-head attention, \ie, using an MLP layer. Essentially, the self-attention mechanism is a kind of directed graphical model \cite{velivckovic2018graph}, while the affinity matrix should be symmetric since nodes sharing the same semantics are supposed to be equal. To perform such transformation, we simply add $S$ and its transpose. The predicted semantic affinity matrix $A\in \mathbb{R}^{hw\times hw}$ is thus denoted as
\begin{equation}
    \label{eq_affinity}
    A = \mathtt{MLP}(S+S^\top).
\end{equation}
Here, we use the matrix transpose operator ${}^\top$ to denote the transpose of each self-attention matrix in tensor $S$.

\noindent\textbf{Pseudo Affinity Label Generation.} To learn the favorable semantic affinity ${A}$, a key step is to derive a reliable pseudo affinity label $Y_{aff}$ as supervision. As shown in Fig.~\ref{fig_wetr}, we derive $Y_{aff}$ from the refined pseudo labels (the refinement module will be introduced later).

\par We first use two background scores $\beta_l$ and $\beta_h$, where $0<\beta_l<\beta_h<1$, to filter the refined pseudo labels to reliable foreground, background, and uncertain regions. Formally, given CAM $M \in \mathbb{R}^{h\times w\times C}$, the pseudo label $Y_p$ is constructed as
\begin{equation}
    \label{eq_ref_pseudo_label}
    Y_p^{i,j}=\left\{
    \begin{aligned}
         & \mathtt{argmax}(M^{i,j,:}), & \text{if $\mathtt{max}(M^{i,j,:})\geq\beta_h$,} \\
         & 0,                          & \text{if $\mathtt{max}(M^{i,j,:})\leq\beta_l$,} \\
         & 255,                        & \text{otherwise,}                               \\
    \end{aligned}
    \right.
\end{equation}
where $0$ and $255$ denote the index of the background class and the ignored regions, respectively. $\mathtt{argmax}(\cdot)$ extracts the semantic class with the maximum activation value.
\par The pseudo affinity label $Y_{aff}\in \mathbb{R}^{hw\times hw}$ is then derived from $Y_p$. Specifically, for $Y_p$, if the pixel $(i, j)$ and $(k,l)$ share the same semantic, we set their affinity as positive; otherwise, their affinity is set as negative. Note that if pixels $(i, j)$ or $(k,l)$ are sampled from the ignored regions, their affinity will also be ignored. Besides, we only consider the situation that pixel $(i,j)$ and $(k,l)$ are in the same local window, and disregard the affinity of distant pixel pairs.

\noindent\textbf{Affinity Loss.} The generated pseudo affinity label $Y_{aff}$ is then used to supervise the predicted affinity $A$. The affinity loss term $\mathcal{L}_{aff}$ is constructed as
\begin{equation}
    \label{eq_loss_aff}
    \small
    \begin{aligned}
        \mathcal{L}_{aff} & = \frac{1}{N^+}\sum_{(ij,kl)\in \mathcal{R}^+}(1-\mathtt{sigmoid}(A^{ij, kl})) \\
                          & + \frac{1}{N^-}\sum_{(ij,kl)\in \mathcal{R}^-}\mathtt{sigmoid}(A^{ij, kl}),
    \end{aligned}
\end{equation}
where $\mathcal{R}^+$ and $\mathcal{R}^-$ denote the set of positive and negative samples in $Y_{aff}$, respectively. $N^+$ and $N^-$ count the number of $\mathcal{R}^+$ and $\mathcal{R}^-$. Intuitively, Eq.~\ref{eq_loss_aff} enforces the network to learn highly confident semantic affinity relations from MHSA. On the other hand, since the affinity prediction $A$ is the linear combination of MHSA, Eq.~\ref{eq_loss_aff} also benefits the learning of self-attention and further helps to discover the integral object regions.

\noindent\textbf{Propagation with Affinity.} The learned reliable semantic affinity could be used to revise the initial CAM. Following \cite{ahn2018learning,ahn2019weakly}, we fulfill this process via random walk \cite{vernaza2017learning}. For the learned semantic affinity matrix $A$, the semantic transition matrix $T$ is derived as
\begin{equation}
    \label{eq_trans_mat}
    T=D^{-1}A^\alpha,\quad \text{with $D^{ii}=\sum_k {A^{ik}}^\alpha$,}
\end{equation}
where $\alpha>1$ is a hyper-parameter to ignore trivial affinity values in $A$, and $D$ is a diagonal matrix to normalize $A$ row-wise. The random walk propagation for the initial CAM $M\in\mathbb{R}^{h\times w\times C}$ is accomplished as
\begin{equation}
    \label{eq_aff_prop}
    M_{aff}=T*\mathtt{vec}(M),
\end{equation}
where $\mathtt{vec}(\cdot)$ vectorizes $M$.
% Unlike \cite{ahn2018learning,ahn2019weakly} perform propagation multiple times, we empirically find single-step propagation operation is preferred in our model. 
This propagation process diffuses the semantic regions with high affinity and dampens the wrongly activated regions so that the activation maps align better with semantic boundaries.

\subsection{Pixel-Adaptive Refinement}\label{sec_lpr}
% \begin{figure}[!tp]
%     \centering
%     \includegraphics[width=0.45\textwidth]{./figures/par.pdf}
%     \caption{Conceptual illustration of PAR. Given a pixel (colored in red), grids in other colors denote its dilated 8-way neighbors.}
%     \label{fig_par}
% \end{figure}
\par As shown in Fig.~\ref{fig_wetr}, the pseudo affinity label $Y_{aff}$ is derived from the initial pseudo labels. However, the initial pseudo labels are typically coarse and locally inconsistent, \ie, neighbor pixels with similar low-level image appearance may not share the same semantic. To ensure the local consistency, \cite{kolesnikov2016seed,zhang2020reliability,zhang2021adaptive} adopt dense CRF \cite{krahenbuhl2011efficient} to refine the initial pseudo labels. However, CRF is not a favorable choice in end-to-end framework since it remarkably slows down the training efficiency. Inspired by \cite{araslanov2020single}, which utilizes the pixel-adaptive convolution \cite{su2019pixel} to extract local RGB information for refinement, we incorporate the RGB and spatial information to define the low-level pairwise affinity and construct our Pixel-Adaptive Refinement module (PAR).
\par Given the input image $I\in\mathbb{R}^{h\times w\times 3}$, for the pixel at position $(i,j)$ and $(k,l)$, the RGB and spatial pairwise terms are defined as:
\begin{equation}
    \label{eq_par_kernel}
    \small
    \kappa^{ij,kl}_{rgb} = -\Bigl(\frac{|I_{ij}-I_{kl}|}{w_1{\sigma^{ij}_{rgb}}}\Bigr)^2,\quad\kappa^{ij,kl}_{pos} = - \Bigl(\frac{|P_{ij}-P_{kl}|}{w_2{\sigma^{ij}_{pos}}}\Bigr)^2,
\end{equation}
where $I_{ij}$ and $P_{ij}$ denote the RGB information and the spatial location of pixel $(i,j)$, respectively. In practice, we use the \texttt{XY} coordinates as the spatial location. In Eq.~\ref{eq_par_kernel}, $\sigma_{rgb}$ and $\sigma_{pos}$ denote the standard deviation of RGB and position difference, respectively. $w_1$ and $w_2$ control the smoothness of $\kappa_{rgb}$ and $\kappa_{pos}$, respectively. The affinity kernel for PAR is then constructed by normalizing $\kappa_{rgb}$ and $\kappa_{pos}$ with \texttt{softmax} and adding them together, \ie,
\begin{equation}
    \label{eq_par_aff}
    \small
    \kappa^{ij,kl} = \frac{\exp(\kappa^{ij,kl}_{rgb})}{\sum_{(x,y)}\exp(\kappa^{ij,xy}_{rgb})}+w_{3}\frac{\exp(\kappa^{ij,kl}_{pos})}{\sum_{(x,y)}\exp(\kappa^{ij,xy}_{pos})},
\end{equation}
where $(x,y)$ is sampled from the neighbor set of $(i, j)$, \ie $\mathcal{N}(i,j)$, and $w_3$ adjusts the importance of the position term. Based on the constructed affinity kernel, we refine both the initial CAM and the propagated CAM. The refinement is conducted for multiple iterations. For CAM ${M}\in\mathbb{R}^{h\times w\times C}$, in iteration $t$, we have
\begin{equation}
    \label{eq_par_update}
    {M}^{i,j,c}_t = \sum_{(k,l)\in\mathcal{N}(i,j)}\kappa^{ij,kl}{M}^{k,l,c}_{t-1}.
\end{equation}
% \par As shown in Fig.~\ref{fig_par}, with our proposed PAR, the mis-activated regions in initial pseudo labels could be effectively dampened. 

For the neighbor pixel sets $\mathcal{N}(\cdot)$, we follow \cite{araslanov2020single} and define it as the 8-way neighbors with multiple dilation rates. Such design ensures the training efficiency, since the dilated neighbors of a given pixel can be easily extracted using $3\times 3$ dilated convolutions.

\subsection{Network Training} \label{sec_loss}
\par As shown in Fig.~\ref{fig_wetr}, our framework consists of three loss terms, \ie, a classification loss $\mathcal{L}_{cls}$, a segmentation loss $\mathcal{L}_{seg}$, and an affinity loss $\mathcal{L}_{aff}$.

For the classification loss, following the common practice, we feed the aggregated features into a classification layer to compute the class probability vector $p_{cls}$, then employ the multi-label soft margin loss as the classification function.
\begin{equation}
    \label{eq_loss_cls}
    \mathcal{L}_{cls} = \frac{1}{C}\sum_{c=1}^C(y^c\log(p_{cls}^c)+(1-{y}^c)\log(1-{p}_{cls}^c)),
\end{equation}
where $C$ is the total number of classes, and $y$ is the ground truth image-level label.
\par For the segmentation loss  $\mathcal{L}_{seg}$, we adopt the commonly-used cross-entropy loss. As shown in Fig.~\ref{fig_wetr}, the supervision for the segmentation branch is the revised label with affinity propagation. In order to obtain better alignment with low-level image appearance, we use the proposed PAR to further refine the propagated labels.
The affinity loss $\mathcal{L}_{aff}$ for affinity learning is previously described in Eq.~\ref{eq_loss_aff}.
\par The overall loss is the weighted sum of $\mathcal{L}_{cls}$, $\mathcal{L}_{aff}$, and $\mathcal{L}_{seg}$. In addition, to further promote the performance, we also employ the regularization loss $\mathcal{L}_{reg}$ used in \cite{tang2018regularized,zhang2021adaptive,zhang2021dynamic,zhang2020reliability}, which ensures the local consistency of the segmentation predictions. The overall loss is finally formulated as
\begin{equation}
    \label{eq_loss}
    \mathcal{L} = \mathcal{L}_{cls} + \lambda_{1}\mathcal{L}_{seg} + \lambda_{2}\mathcal{L}_{aff} + \lambda_{3}\mathcal{L}_{reg},
\end{equation}
where $\lambda_1,\lambda_2$, and $\lambda_3$ balance the contributions of different losses.

\section{Experiments}
\subsection{Setup}
\noindent\textbf{Datasets.} We conduct experiments on \textit{PASCAL VOC 2012} and \textit{MS COCO 2014} datasets. {PASCAL VOC 2012} dataset \cite{everingham2010pascal} contains 21 semantic classes (including the \textit{background} class). This dataset is usually augmented with the SBD dataset \cite{hariharan2011semantic}. The augmented dataset includes 10,582, 1,449, and 1,464 images for training, validation, and testing, respectively. {MS COCO 2014} dataset \cite{lin2014microsoft} contains 81 classes and includes 82,081 images for training and 40,137 images for validation. The images in $train$ sets of {PASCAL VOC} and {MS COCO} are annotated with image-level labels only. By default, we report mean Intersection-Over-Union (mIoU) as the evaluation criteria.

\noindent\textbf{Network Configuration.} For the Transformer backbone, we use the Mix Transformer (MiT) proposed in Segformer \cite{xie2021segformer}, which is a more friendly backbone for image segmentation tasks than the vanilla ViT \cite{zheng2021rethinking}. In brief, MiT uses overlapped patch embedding to keep local consistency, spatial-reductive self-attention to accelerate computation, and \texttt{FFN} with convolutions to safely replace position embedding. For the segmentation decoder, we use the MLP decoder head \cite{xie2021segformer}, which fuses multi-level feature maps for prediction with simple MLP layers. The backbone parameters are initialized with ImageNet-1k \cite{deng2009imagenet} pre-trained weights, while other parameters are randomly initialized.

\noindent\textbf{Implementation Details.} We use an AdamW optimizer \cite{loshchilov2019decoupled} to train our network. For the backbone parameters, the initial learning rate is set as $6\times 10^{-5}$ and decays every iteration with a polynomial scheduler. The learning rate for other parameters is 10 times the learning rate of backbone parameters. The weight decay factor is set as $0.01$.
For data augmentation, random rescaling with a range of $[0.5, 2.0]$, random horizontally flipping, and random cropping with a cropping size of $512\times 512$ are adopted. The batch size is set as 8.
For the experiments on the PASCAL VOC dataset, we train the network for 20,000 iterations. To ensure the initial pseudo labels are favorable, we warm-up the classification branch for 2,000 iterations and the affinity branch for the next 4,000 iterations. For experiments on the MS COCO dataset, the number of total iterations is 80,000. Accordingly, the number of warm-up iteration for the classification and affinity branch are 5,000 and 15,000, respectively.
\par The default hyper-parameters are set as follows. For pseudo label generation, the background thresholds $(\beta_h, \beta_l)$ are $(0.55, 0.35)$. In PAR, same as \cite{araslanov2020single}, the dilation rates for extracting neighbor pixels are $[1, 2, 4, 8, 12, 24]$. We set the weight factors $(w_1, w_2, w_3)$ as $(0.3, 0.3, 0.01)$. When computing the affinity loss, the radius of the local window to ignore distant affinity pairs is set as $8$. In Eq.~\ref{eq_trans_mat}, we set the power factor $\alpha$ as 2. The weight factors in Eq.~\ref{eq_loss} are 0.1, 0.1, and 0.01, respectively. The detailed investigation of the hyper-parameters is reported in the supplementary material.

\begin{table}[!t]
    \caption{Impact of \texttt{top-k} pooling with different top percentages on CAM. The results are evaluated on the PASCAL VOC $train$ and $val$ set and reported in mIoU(\%).}
    \label{tab_topk_pooling}
    \centering
    \small
    \setlength{\tabcolsep}{0.021\textwidth}
    \begin{tabular}{l|cccccc}
        \toprule
                & \texttt{gap} & 50\% & 25\% & 10\% & \texttt{gmp}  \\ \midrule
        $train$ & 30.7         & 34.5 & 39.6 & 43.5 & \textbf{48.2} \\
        $val$   & 31.1         & 34.8 & 39.7 & 43.6 & \textbf{48.3} \\ \bottomrule
    \end{tabular}
    \vspace{-4mm}
\end{table}

\subsection{Initial Pseudo Label Generation.}
\par In this work, we use the popular CAM to generate the initial pseudo labels. Empirically, for a CNN-based classification network, the choice of pooling method notably affects the quality of CAM. Specifically, global max-pooling (\texttt{gmp}) tends to underestimate the object size, while global average-pooling (\texttt{gap}) typically overestimates the object regions \cite{kolesnikov2016seed,zhou2016learning}. Here, we investigate the favorable pooling method for Transformer-based classification network. We first generalize \texttt{gmp} and \texttt{gap} with \texttt{top-k} pooling, \ie, averaging the top $k$\% values in each feature map. In this situation, \texttt{gmp} and \texttt{gap} are two special cases of \texttt{top-k} pooling, \ie, top-100\% and top-1 pooling. We present the impact of \texttt{top-k} pooling with different $k$ in Tab.~\ref{tab_topk_pooling}. Tab.~\ref{tab_topk_pooling} shows that in our framework, for the Transformer-based classification network, using \texttt{gmp} for feature aggregation helps to generate CAM with favorable performance, which is owing to the capacity of global modeling of self-attention.

\subsection{Ablation Study and Analysis}
\begin{table}[!tbp]
    \caption{Ablation studies of our proposed method on PASCAL VOC $val$ set.}
    \label{tab_ablation}
    \small
    \centering
    \setlength{\tabcolsep}{0.017\textwidth}
    \begin{tabular}{l|cccc|c}
        \toprule
        Method                & PAR    & AFA    & $\mathcal{L}_{reg}$ & CRF    & $val$ \\ \midrule
        Our Baseline          &        &        &                     &        & 46.7  \\ \midrule
        \multirow{4}{*}{Ours} & \cmark &        &                     &        & 56.2  \\
                              & \cmark & \cmark &                     &        & 62.6  \\
                              & \cmark & \cmark & \cmark              &        & 63.8  \\
                              & \cmark & \cmark & \cmark              & \cmark & 66.0  \\ \bottomrule
    \end{tabular}
\end{table}
\par The quantitative results of ablation analysis are reported in Tab.~\ref{tab_ablation}. Tab.~\ref{tab_ablation} shows that our baseline model based on Transformers achieves 46.7\% mIoU on the PASCAL VOC $val$ set. The proposed PAR and AFA further significantly improve the mIoU to 56.2\% and 62.6\%, respectively. With the auxiliary regularization loss $\mathcal{L}_{reg}$, the proposed framework achieves 63.8\% mIoU. The CRF post-processing brings further 2.2\% mIoU improvements, promoting the final performance to 66.0\% mIoU. In short, the quantitative results in Tab.~\ref{tab_ablation} demonstrate our proposed modules are remarkably effective.
\begin{figure*}[htp]
    \centering
    \includegraphics[width=0.95\textwidth]{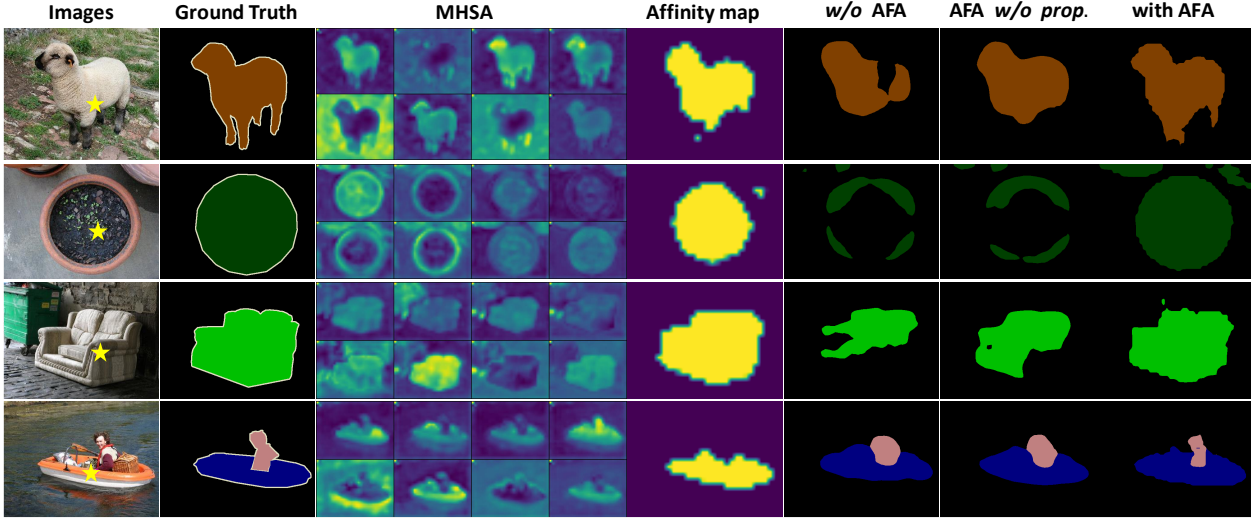}
    \caption{Visualization of the MHSA maps, learned affinity maps, and generated pseudo labels for segmentation. "$\bigstar$" denotes the query point to visualize the attention and affinity maps.}
    \vspace{-4mm}
    \label{fig_vis_afa}
\end{figure*}

\setlength{\columnsep}{10pt}%
\begin{table}[t]
    \centering
    \small
    \caption{Evaluation of the pseudo labels for segmentation.}
    \label{tab_aff_prop}
    \setlength{\tabcolsep}{0.017\textwidth}
    \begin{tabular}{l|l|cc}
        \toprule
                                          &                            & $train$       & $val$         \\ \midrule
        PSA \cite{ahn2018learning}        &                            & 59.7          & --            \\
        IRN  \cite{ahn2019weakly}         &                            & 66.5          & --            \\
        1Stage \cite{araslanov2020single} &                            & 66.9          & 65.3          \\ \midrule
        \multirow{4}{*}{Ours}             & \textit{w/o} AFA           & 54.4          & 54.2          \\
                                          & AFA (\textit{w/o} $prop.$) & 66.3          & 64.4          \\
                                          & AFA ($prop.$ with MHSA)    & 58.3          & 55.9          \\
                                          & AFA                        & \textbf{68.7} & \textbf{66.5} \\ \bottomrule
    \end{tabular}
    \vspace{-4mm}
\end{table}%\vspace{-5mm}

\noindent\textbf{AFA.}
The motivation of AFA is to learn reliable semantic affinity from MHSA and revise the pseudo labels with the learned affinity. In Fig.~\ref{fig_vis_afa}, we present some example images of the self-attention maps (extracted from the last Transformer block) and the learned affinity maps. Fig.~\ref{fig_vis_afa} shows that our AFA could effectively learn reliable semantic affinity from the inaccurate MHSA. The affinity loss in the AFA module also encourages the MHSA to model the semantic relations well. In Fig.~\ref{fig_vis_afa}, we also present the pseudo labels generated from our model without AFA module (\textit{w/o} AFA), with AFA module but no random walk propagation (AFA \textit{w/o} $prop.$) and with full AFA module. For the generated pseudo labels, the AFA module brings notable visual improvements. The affinity propagation process further diffuses the regions with high semantic affinity and dampens the regions with low affinity.
\par In Tab.~\ref{tab_aff_prop}, we report the quantitative results of the generated pseudo labels on PASCAL VOC $train$ and $val$ set. We also report the results of performing random walk propagation with the average vanilla MHSA as semantic affinity (AFA $prop.$ with MHSA). The results show that the affinity learning loss in the AFA module remarkably improves the accuracy of the pseudo labels (from 54.4\% mIoU to 66.3\% mIoU on the $train$ set). The propagation process could further promote the reliability of pseudo labels, which harvests the performance gains in Tab.~\ref{tab_ablation}. It is also noted that propagation with the naive MHSA significantly reduces the accuracy, demonstrating our motivation and the effectiveness of the AFA module.

\begin{figure}[!tp]
    \centering
    \includegraphics[width=0.35\textwidth]{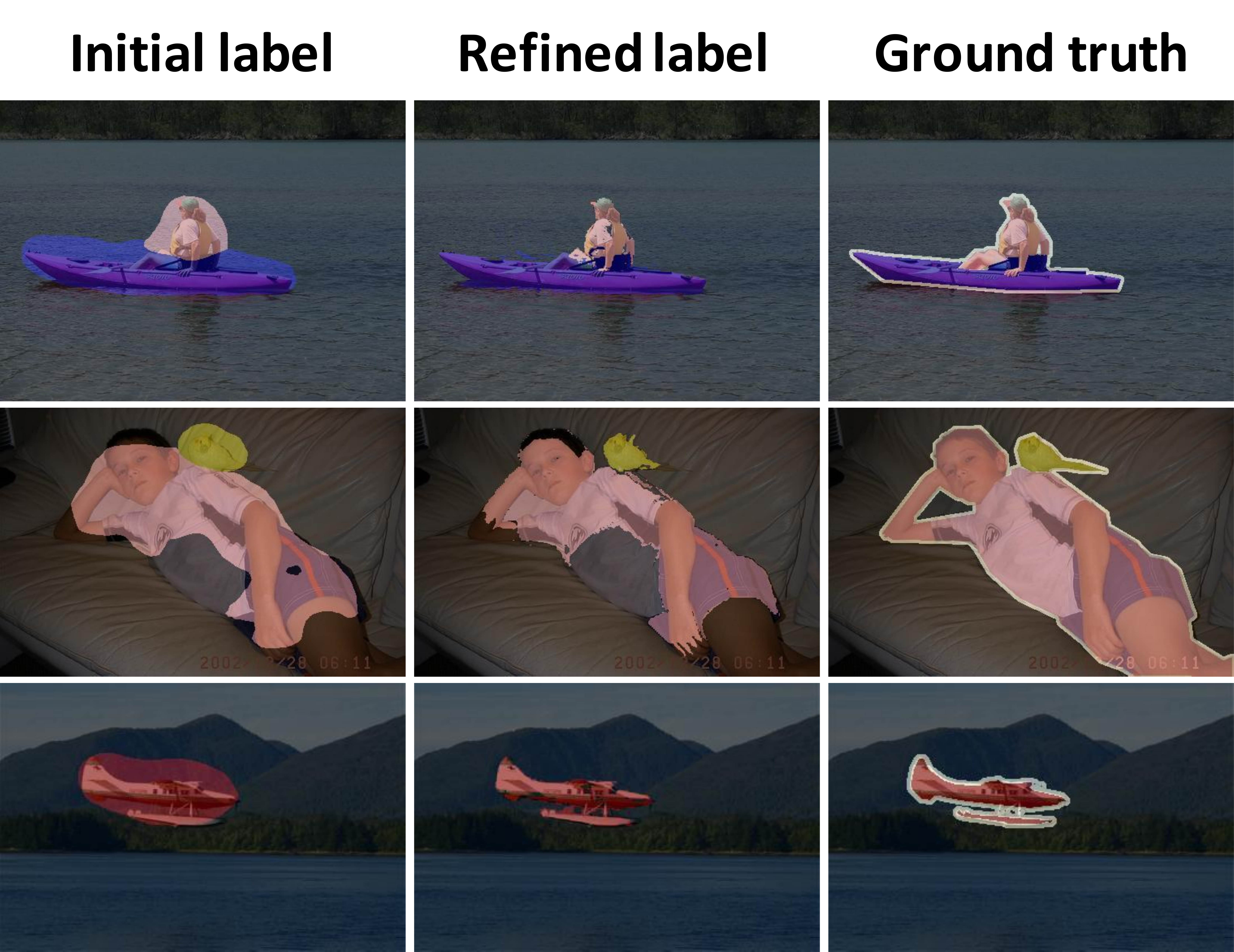}
    \caption{Examples of PAR's improvements on the pseudo labels. The pseudo labels are generated with CAM and Transformer baseline.}
    \vspace{-2mm}
    \label{fig_vis_par}
\end{figure}

\noindent\textbf{PAR.} The proposed PAR aims at refining the initial pseudo labels with low-level image appearance and position information. In Fig.~\ref{fig_vis_par}, we present the qualitative improvements of PAR. Fig.~\ref{fig_vis_par} shows PAR effectively dampens the falsely activated regions, enforcing better alignment with the low-level boundaries.
\par Quantitatively, as shown in Tab.~\ref{tab_par}, our PAR improves the CAM (generated with Transformer baseline) from 48.2\% to 52.9\%, which outperforms PAMR \cite{araslanov2020single}, which is also based on the dilated pixel-adaptive convolution to incorporate local image appearance information. Tab.~\ref{tab_par} also demonstrates the position kernel $\kappa_{pos}$ in PAR is beneficial for refining CAM. More investigation details on PAR are presented in the supplementary material.

\setlength{\columnsep}{8pt}%
\begin{table}[!tp]
    \caption{Comparison of refinement methods for CAM.}
    \label{tab_par}
    \centering
    \small
    \setlength{\tabcolsep}{0.035\textwidth}
    \begin{tabular}{l|cc|c}
        \toprule
                                        & {$\kappa_{rgb}$} & {$\kappa_{pos}$} & {$train$}     \\ \midrule
        CAM                             &                  &                  & 48.2          \\
        PAMR \cite{araslanov2020single} & \cmark           &                  & 51.4          \\ \midrule
        \multirow{2}{*}{PAR}            & \cmark           &                  & 51.7          \\
                                        & \cmark           & \cmark           & \textbf{52.9} \\ \bottomrule
    \end{tabular}
    \vspace{-4mm}
\end{table}

\begin{table}[!tp]

    \centering
    \caption{Semantic segmentation results on PASCAL VOC 2012 dataset. $Sup.$ denotes supervision type.  $\mathcal{F}$: full supervision; $\mathcal{I}$: image-level labels; $\mathcal{S}$: saliency maps.  $\dagger$ denotes our implementation.}
    % \vspace{-2mm}
    \small
    \label{tab_voc_seg}
    \begin{tabular}{l|c|l|cc}
        \toprule
        {Method}                                                    & {$Sup.$}                                   & Backbone & {$val$}       & {$test$}      \\\midrule
        \multicolumn{5}{l}{\cellcolor[HTML]{ffffff}\textbf{\textit{Fully-supervised models}}.}                                                              \\
        DeepLab \cite{chen2017deeplab}                              & \multirow{3}{*}{$\mathcal{F}$}             & R101     & 77.6          & 79.7          \\
        WideResNet38   \cite{wu2019wider}                           &                                            & WR38     & 80.8          & 82.5          \\
        Segformer  $^\dagger$   \cite{xie2021segformer}                         &                                            & MiT-B1   & 78.7          & --            \\ \midrule
        \multicolumn{5}{l}{\cellcolor[HTML]{ffffff}\textbf{\textit{Multi-Stage weakly-supervised models}}.}                                                 \\
        OAA+ \cite{jiang2019integral} \tiny ICCV'2019               & \multirow{5}{*}{$\mathcal{I}+\mathcal{S}$} & R101     & 65.2          & 66.4          \\
        MCIS \cite{sun2020mining} \tiny ECCV'2020                   &                                            & R101     & 66.2          & 66.9          \\
        AuxSegNet \cite{xu2021leveraging} \tiny ICCV'2021           &                                            & WR38     & 69.0          & 68.6          \\
        NSROM \cite{yao2021non} \tiny CVPR'2021                     &                                            & R101     & 70.4          & 70.2          \\
        EPS \cite{lee2021railroad} \tiny CVPR'2021                  &                                            & R101     & \textbf{70.9} & \textbf{70.8} \\\midrule
        SEAM \cite{wang2020self}  \tiny CVPR'2020                   & \multirow{6}{*}{$\mathcal{I}$}             & WR38     & 64.5          & 65.7          \\
        SC-CAM \cite{chang2020weakly} \tiny CVPR'2020               &                                            & R101     & 66.1          & 65.9          \\
        CDA \cite{su2021context} \tiny ICCV'2021                    &                                            & WR38     & 66.1          & 66.8          \\
        AdvCAM \cite{lee2021anti} \tiny CVPR'2021                   &                                            & R101     & 68.1          & 68.0          \\
        CPN \cite{zhang2021complementary} \tiny ICCV'2021           &                                            & R101     & 67.8          & 68.5          \\
        RIB \cite{lee2021reducing} \tiny NeurIPS'2021               &                                            & R101     & \textbf{68.3} & \textbf{68.6} \\ \midrule
        \multicolumn{5}{l}{\cellcolor[HTML]{ffffff}\textbf{\textit{End-to-End weakly-supervised models}}.}                                                  \\
        EM \cite{papandreou2015weakly}  \tiny ICCV'2015             & \multirow{7}{*}{$\mathcal{I}$}             & VGG16    & 38.2          & 39.6          \\
        MIL \cite{pinheiro2015image}  \tiny CVPR'2015               &                                            & --       & 42.0          & 40.6          \\
        CRF-RNN \cite{roy2017combining}  \tiny CVPR'2017            &                                            & VGG16    & 52.8          & 53.7          \\
        RRM \cite{zhang2020reliability}  \tiny AAAI'2020            &                                            & WR38     & 62.6          & 62.9          \\
        RRM $^\dagger$ \cite{zhang2020reliability}  \tiny AAAI'2020 &                                            & MiT-B1   & 63.5          & --            \\
        1Stage \cite{araslanov2020single} \tiny CVPR'2020           &                                            & WR38     & 62.7          & 64.3          \\
        AA\&LR \cite{zhang2021adaptive} \tiny ACM MM'2021           &                                            & WR38     & 63.9          & 64.8          \\
        \rowcolor [HTML]{eaeaea}
        \textbf{Ours}                                               &                                            & MiT-B1   & \textbf{66.0} & \textbf{66.3} \\ \bottomrule
    \end{tabular}
    \vspace{-2mm}
\end{table}

\subsection{Comparison to State-of-the-art}
\noindent\textbf{PASCAL VOC 2012.} We report the semantic segmentation performance on PASCAL VOC 2012 $val$ and $test$ set in Tab.~\ref{tab_voc_seg}. R101 and WR38 denote the method uses ResNet101 \cite{he2016deep} and WideResNet38 \cite{wu2019wider} as backbone, respectively. Tab.~\ref{tab_voc_seg} shows that the proposed model clearly surpasses previous state-of-the-art end-to-end methods. Our method achieves 83.8\% of its fully-supervised counterpart, \ie, Segformer \cite{xie2021segformer}, while 1Stage \cite{araslanov2020single} and AA\&LR \cite{zhang2021adaptive} only achieve 77.6\% and 79.1\% of WideResNet38, respectively. Our method is also competitive with some recent multi-stage WSSS methods, such as OAA+ \cite{jiang2019integral}, SEAM \cite{wang2020self}, SC-CAM \cite{chang2020weakly}, and CDA \cite{su2021context}. It's also noted that our method also outperforms RRM with MiT-B1 as backbone, which demonstrates the efficay of the proposed AFA and PAR.
% In specific, our method achieves 65.1 and 65.2 mIoU on the VOC $val$ and $test$ set, respectively. On the $val$ set, our method surpassed AA\&LR \cite{zhang2021adaptive} by 1.2\% and 1Stage \cite{araslanov2020single} by 2.4\%, respectively.

\noindent\textbf{MS COCO 2014.} We present the semantic segmentation performance on the challenging MS COCO 2014 dataset in Tab.~\ref{tab_coco_seg}. Our end-to-end method could achieve 38.9\% mIoU on the $val$ set, which remarkably outperforms most recent multi-stage methods (except RIB \cite{lee2021reducing}).

\begin{table}[!t]
    \centering
    \caption{Semantic segmentation results on MS COCO dataset. }
    \small
    \label{tab_coco_seg}
    \setlength{\tabcolsep}{0.018\textwidth}
    \begin{tabular}{l|c|l|c}
        \toprule
        {Method}                                          & {$Sup.$}                                   & Backbone & {$val$}       \\\midrule

        \multicolumn{4}{l}{\cellcolor[HTML]{ffffff}\textbf{\textit{Multi-Stage weakly-supervised models}}.}                       \\
        EPS \cite{lee2021railroad} \tiny CVPR'2021        & \multirow{2}{*}{$\mathcal{I}+\mathcal{S}$} & R101     & \textbf{35.7} \\
        AuxSegNet \cite{xu2021leveraging} \tiny ICCV'2021 &                                            & WR38     & 33.9          \\ \midrule
        SEAM \cite{wang2020self} \tiny CVPR'2020          & \multirow{5}{*}{$\mathcal{I}$}             & WR38     & 31.9          \\
        CONTA \cite{zhang2020causal} \tiny NeurIPS'2020   &                                            & WR38     & 32.8          \\
        CDA \cite{su2021context} \tiny ICCV'2021          &                                            & WR38     & 31.7          \\
        CGNet \cite{kweon2021unlocking} \tiny ICCV'2021   &                                            & WR38     & 36.4          \\
        RIB \cite{lee2021reducing} \tiny NeurIPS'2021     &                                            & R101     & \textbf{43.8} \\\midrule
        \multicolumn{4}{l}{\cellcolor[HTML]{ffffff}\textbf{\textit{End-to-End weakly-supervised models}}.}                        \\
        \rowcolor [HTML]{eeeeee}
        \textbf{Ours}                                     &                                            & MiT-B1   & 38.0          \\
        \rowcolor [HTML]{eeeeee}
        \textbf{Ours + CRF}                               & \multirow{-2}{*}{$\mathcal{I}$}            & MiT-B1   & \textbf{38.9} \\\bottomrule
    \end{tabular}
    \vspace{-2mm}

\end{table}

\begin{figure}[!tp]
    \centering
    \includegraphics[width=0.48\textwidth]{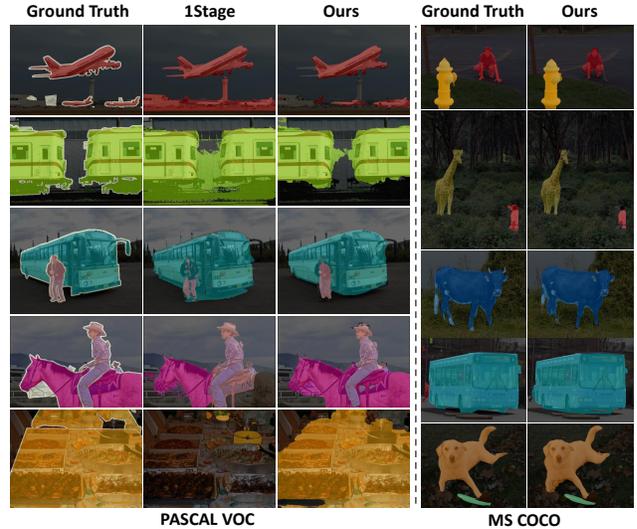}
    \vspace{-4mm}
    \caption{Segmentation results of 1Stage \cite{araslanov2020single} and our method on VOC and COCO $val$ set.}
    \label{fig_vis_segs}
    \vspace{-4mm}
\end{figure}

\noindent\textbf{Qualitative Results.} In Fig.~\ref{fig_vis_segs}, we present the qualitative results of our method on the PASCAL VOC and MS COCO $val$ set. On the PASCAL VOC dataset, visually, our method could outperform 1Stage \cite{araslanov2020single} and produce segmentation results that align finely with object boundaries. The qualitative results on MS COCO dataset are also comparable with the ground truth.

\section{Conclusion}
In this work, we explore the intrinsic virtue of Transformer architecture for WSSSS tasks. Specifically, we use a Transformer-based backbone to generate CAM as the initial pseudo labels, avoiding the inherent flaw of CNN. Besides, we note the consistency between the MHSA and semantic affinity, and thus propose the AFA module. AFA derives reliable affinity labels from pseudo labels, imposes the affinity labels to supervise the MHSA, and produces reliable affinity predictions. The learned affinity is used to revise the initial pseudo labels via random walk propagation. On PASCAL VOC and MS COCO datasets, our method achieves new state-of-the-art performance for end-to-end WSSS. In a broader view, the proposed method also shows a novel perspective for vision transformers, \ie guiding the self-attention with semantic relation to ensure better feature aggregation.
\vspace{-2mm}
\section*{Acknowledgments} \noindent This work was supported in part by the National Natural Science Foundation of China under Grants 62141112, 41871243, and 62002090, the Science and Technology Major Project of Hubei Province (Next-Generation AI Technologies) under Grant 2019AEA170, the Major Science and Technology Innovation 2030 "New Generation Artificial Intelligence" key project (No. 2021ZD0111700). Dr. Baosheng Yu is supported by ARC project FL-170100117.

{\small
\bibliographystyle{ieee_fullname}
\bibliography{afa.bbl}
}

\newpage

\section{More Technical Details}
\subsection{Details of Backbone}
\begin{figure}[htp]
    \centering
    \includegraphics[width=0.47\textwidth]{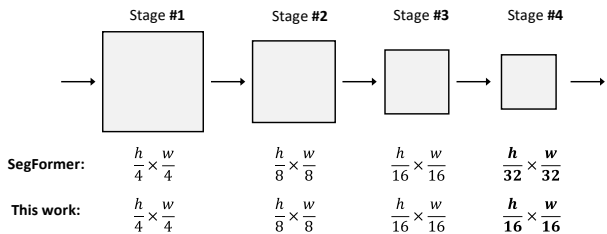}
    \caption{The size of feature maps of different stags.}
    \label{fig_backbone}
\end{figure}

We use the MiT-B1 proposed in SegFormer \cite{xie2021segformer} as the backbone, which is a more friendly backbone for image segmentation tasks than the vanilla ViT \cite{dosovitskiy2020image}. SegFormer uses Overlapped Patch Merging layers with different strides to produce multi-scale feature maps. As shown in Fig.~\ref{fig_backbone}, in SegFormer, the feature of Stage \#4 is $\frac{h}{32}\times\frac{w}{32}$. To obtain the initial pseudo labels (CAM) with higher resolution, we change the stride of the last patch merging layer from 2 to 1, increasing resolution of the feature maps to the size of $\frac{h}{16}\times\frac{w}{16}$.
\par In practice, to produce the semantic affinity prediction, we use the multi-head self-attention (MHSA) matrices extracted from the last stage, which could capture the high-level semantic information. The MHSA matrices are concatenated to form $S\in\mathbb{R}^{\frac{hw}{256}\times\frac{hw}{256}\times nk}$ and prediction the semantic affinity, where $n$ and $k$ are the number of Transformer blocks and heads in each block, respectively.

\subsection{Mask for Affinity Loss}
\begin{figure}[htp]
    \centering
    \includegraphics[width=0.45\textwidth]{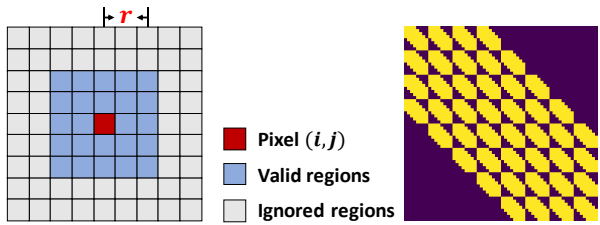}
    \caption{\textit{Left}: Illustration of the valid pixel pairs. \textit{Right}: Example mask for computing the affinity loss.}
    \label{fig_aff_mask}
\end{figure}

Inspired by \cite{ahn2019weakly,ahn2018learning}, when computing affinity loss, we only consider the situation that pixel pairs are in the same local window with the radius of $r$, and disregard their affinity if the distance is too far. Specifically, given a pixel $(i,j)$, if pixel $(k,l)$ is the same window with $(i,j)$, their affinity is computed; otherwise, their affinity is ignored. Unlike \cite{ahn2019weakly,ahn2018learning}, which extract pixel pairs when computing affinity loss, we efficiently implemented by applying a mask. The conceptual illustration of this strategy and an example mask is presented in Fig.~\ref{fig_aff_mask}.

\section{More Experimental Results}
\subsection{Hyper-parameters}
\noindent\textbf{Affinity from Attention.} In Tab.~\ref{tab_radius}, we present segmentation results on the PASCAL VOC $val$ set with different radius $r$ of the local window size when computing the affinity loss. Intuitively, a small $r$ can not provide enough affinity pairs while a large $r$ may not ensure the reliability of distant affinity pairs. As shown in Tab.~\ref{tab_radius}, $r=8$ is a proper choice.
\begin{table}[h]
    \centering
    \small
    \setlength{\tabcolsep}{0.55em}%
    \caption{Impact of the radius $r$ when computing the affinity loss. The results are evaluated on the $val$ set of PASCAL VOC 2012.}
    \label{tab_radius}
    \setlength{\tabcolsep}{0.021\textwidth}
    \begin{tabular}{c|ccccc}
        \toprule
        radius $r$ & 2    & 4    & \cellcolor [HTML]{e0e0e0} 8            & 12   & 16   \\ \midrule
        $val$      & 62.4 & 62.7 & \cellcolor [HTML]{e0e0e0}\textbf{63.8} & 61.5 & 59.4 \\ \bottomrule
    \end{tabular}
\end{table}

\noindent\textbf{Pixel-Adaptive Refinement.} In Tab.~\ref{tab_par_1}, we report the impact of different configurations of the proposed Pixel-Adaptive Refinement, including the dilation rates, position kernel, and the number of iteration. Tab.~\ref{tab_par_1} shows that for the same dilation rates, our PAR remarkably outperforms PAMR \cite{araslanov2020single}, demonstrating the necessity of the position kernel.
\begin{table}[ht]
    \small
    \centering
    \setlength{\tabcolsep}{0.55em}%
    \caption{Ablation of the dilation rates, position kernel and number of iteration of the proposed PAR. The results are evaluated on the $train$ set of PASCAL VOC 2012 in mIoU (\%).}
    \label{tab_par_1}
    \begin{tabular}{l|llllll|l|l|l}
        \toprule
                                       & \multicolumn{6}{c|}{Dilations} & \multirow{2}{*}{$\kappa_{pos}$} & \multirow{2}{*}{Iter} & \multirow{2}{*}{$train$}                                                  \\\cmidrule{2-7}
                                       & 1                              & 2                               & 4                     & 8                        & 12     & 24     &        &    &                \\ \midrule
        CAM                            &                                &                                 &                       &                          &        &        &        &    & 48.2           \\ \midrule
        PAMR\cite{araslanov2020single} & \cmark                         & \cmark                          & \cmark                & \cmark                   & \cmark & \cmark &        &    & 51.4           \\
        CRF                            &                                &                                 &                       &                          &        &        &        &    & \textbf{54.5}  \\ \midrule
        \multirow{6}{*}{PAR}           & \cmark                         & \cmark                          & \cmark                &                          &        &        & \cmark & 15 & 48.8           \\
                                       & \cmark                         & \cmark                          & \cmark                & \cmark                   &        &        & \cmark & 15 & 49.9           \\
                                       & \cmark                         & \cmark                          & \cmark                & \cmark                   & \cmark &        & \cmark & 15 & 51.3           \\
                                       & \cmark                         & \cmark                          & \cmark                & \cmark                   & \cmark & \cmark &        & 15 & 51.5           \\
        \rowcolor [HTML]{e0e0e0}
                                       & \cmark                         & \cmark                          & \cmark                & \cmark                   & \cmark & \cmark & \cmark & 15 & \textbf{52.9}  \\
                                       & \cmark                         & \cmark                          & \cmark                & \cmark                   & \cmark & \cmark & \cmark & 20 & \textbf{52.9 } \\ \bottomrule
    \end{tabular}
\end{table}

\par Tab.~\ref{tab_par_2} presents the impact of the weights factors of PAR. For simplicity, we set $w_1=w_2$. Tab.~\ref{tab_par_2} shows $w_1=0.3, w_2=0.3, w_3=0.01$ is a favorable choice.
\begin{table}[]
    \centering
    \small
    \caption{Ablation of weight factors of the proposed PAR. The results are evaluated on the $train$ set of PASCAL VOC 2012.}
    \label{tab_par_2}
    \setlength{\tabcolsep}{0.02\textwidth}
    \begin{tabular}{c|c|cccc}
        \toprule
        \multicolumn{2}{c|}{\multirow{2}{*}{}} & \multicolumn{4}{c}{$w_3$}                                                                                            \\ \cmidrule{3-6}
        \multicolumn{2}{c|}{}                  & 0.005                        & \cellcolor [HTML]{e0e0e0} 0.01 & 0.02                                   & 0.03        \\ \midrule
        \multirow{4}{*}{$w_1$ \& $w_2$}        & 0.1                          & 51.9                           & 51.7                                   & 50.1 & --   \\
                                               & \cellcolor [HTML]{e0e0e0}0.3 & 52.8                           & \cellcolor [HTML]{e0e0e0}\textbf{52.9} & 51.4 & 48.4 \\
                                               & 0.5                          & 51.9                           & 52.5                                   & 51.3 & 48.3 \\
                                               & 0.7                          & --                             & 51.6                                   & 50.9 & 48.0 \\ \bottomrule
    \end{tabular}
\end{table}

\noindent\textbf{Weight Factors.} We present the segmentation results on the PASCAL VOC $val$ set with different weight factors of loss terms in Tab.~\ref{tab_loss_w}. $\lambda_1=0.1, \lambda_2=0.2, \lambda_2=0.01$ is a preferred choice for our framework.
\begin{table}[]
    \centering
    \caption{Impact of the weights of loss terms. The results are evaluated on the $val$ set of PASCAL VOC 2012.}
    \label{tab_loss_w}
    \small
    \setlength{\tabcolsep}{0.027\textwidth}
    \begin{tabular}{l|ccc|c}
        \toprule
                & $\lambda_1$ & $\lambda_2$ & $\lambda_3$ & $val$         \\ \midrule
        \rowcolor[HTML]{e0e0e0}
        Default & 0.1         & 0.1         & 0.01        & \textbf{63.8} \\ \midrule
                & 0.05        &             &             & 62.8          \\
                & 0.2         &             &             & 61.6          \\
                & 0.5         &             &             & 57.8          \\ \cmidrule{2-5}
                &             & 0.05        &             & 63.4          \\
                &             & 0.2         &             & 61.7          \\
                &             & 0.5         &             & 58.7          \\ \cmidrule{2-5}
                &             &             & 0.005       & 62.4          \\
                &             &             & 0.02        & 62.3          \\
                &             &             & 0.05        & 61.5          \\ \bottomrule
    \end{tabular}
\end{table}

\noindent\textbf{Background Scores} We investigate the impact of the background scores $(\beta_l, \beta_h)$ to filter the pseudo labels to the reliable foreground, background, and uncertain regions. Intuitively, large $\beta_h$ and small $\beta_l$ could produce more reliable pseudo labels but reduce the number of valid labels. On the contrary, small $\beta_h$ and large $\beta_l$ will introduce noise to the pseudo labels. Note that the average value of $\beta_h$ and $\beta_l$ is always 0.45, which is the preferred background score for generated CAM in our preliminary experiments.
\begin{table}[]
    \centering
    \caption{Impact of the background scores $\beta_h, \beta_l$. The results are evaluated on the $val$ set of PASCAL VOC 2012.}
    \small
    \setlength{\tabcolsep}{0.04\textwidth}
    \begin{tabular}{l|cc|c}
        \toprule
                & $\beta_h$ & $\beta_l$ & $val$         \\ \midrule
                & 0.65      & 0.25      & 60.7          \\
                & 0.6       & 0.3       & 62.5          \\
        \rowcolor[HTML]{e0e0e0}
        Default & 0.55      & 0.35      & \textbf{63.8} \\
                & 0.5       & 0.4       & 62.9          \\
                & 0.45      & 0.45      & 60.5          \\  \bottomrule
    \end{tabular}
\end{table}

\begin{table}[!tp]
    \small
    \centering
    \setlength{\tabcolsep}{0.55em}%
    \caption{Evaluation and comparison of the semantic segmentation results in mIoU on the $val$ set.}
    \label{tab_miou}
    \setlength{\tabcolsep}{0.012\textwidth}
    \begin{tabular}{l|ccc|c}
        \toprule
                         & {RRM}\cite{zhang2020reliability} & {1Stage} \cite{araslanov2020single} & {AA\&LR} \cite{zhang2021adaptive} & \textbf{Ours}  \\ \midrule
        \textbf{bkg}     & 87.9                             & 88.7                                & 88.4                              & \textbf{89.9}  \\
        \textbf{aero}    & 75.9                             & 70.4                                & 76.3                              & \textbf{79.5}  \\
        \textbf{bicycle} & 31.7                             & \textbf{35.1}                       & 33.8                              & 31.2           \\
        \textbf{bird}    & 78.3                             & 75.7                                & 79.9                              & \textbf{80.7}  \\
        \textbf{boat}    & 54.6                             & 51.9                                & 34.2                              & \textbf{67.2}  \\
        \textbf{bottle}  & 62.2                             & 65.8                                & \textbf{68.2}                     & 61.9           \\
        \textbf{bus}     & 80.5                             & 71.9                                & 75.8                              & \textbf{81.4}  \\
        \textbf{car}     & 73.7                             & 64.2                                & \textbf{74.8}                     & 65.4           \\
        \textbf{cat}     & 71.2                             & 81.1                                & 82.0                              & \textbf{82.3}  \\
        \textbf{chair}   & 30.5                             & 30.8                                & \textbf{31.8}                     & 28.7           \\
        \textbf{cow}     & 67.4                             & 73.3                                & 68.7                              & \textbf{83.4}  \\
        \textbf{table}   & 40.9                             & 28.1                                & \textbf{47.4}                     & 41.6           \\
        \textbf{dog}     & 71.8                             & 81.6                                & 79.1                              & \textbf{82.2 } \\
        \textbf{horse}   & 66.2                             & 69.1                                & 68.5                              & \textbf{75.9}  \\
        \textbf{motor}   & 70.3                             & 62.6                                & \textbf{71.4}                     & 70.2           \\
        \textbf{person}  & 72.6                             & 74.8                                & \textbf{80.0}                     & 69.4           \\
        \textbf{plant}   & 49.0                             & 48.6                                & 50.3                              & \textbf{53.0}  \\
        \textbf{sheep}   & 70.7                             & 71.0                                & 76.5                              & \textbf{85.9}  \\
        \textbf{sofa}    & 38.4                             & 40.1                                & 43.0                              & \textbf{44.1}  \\
        \textbf{train}   & 62.7                             & \textbf{68.5}                       & 55.5                              & 64.2           \\
        \textbf{tv}      & 58.4                             & \textbf{64.3}                       & 58.5                              & 50.9           \\ \midrule
        \textbf{mIOU}    & 62.6                             & 62.7                                & 63.9                              & \textbf{66.0}  \\ \bottomrule
    \end{tabular}
\end{table}

\subsection{More Quantitative Results}
We present the per-category segmentation results on PASCAL VOC $val$ set in Tab~\ref{tab_miou}. Our method achieves the best results for most categories. The results on $test$ set are available at the official PASCAL VOC evaluation website\footnote{\url{http://host.robots.ox.ac.uk:8080/anonymous/GHJIIH.html}}.

\subsection{More Qualitative Results}
We present more qualitative results as follows.
\vspace{5cm}
\begin{figure*}[!tp]
    \centering
    \includegraphics[width=0.9\textwidth]{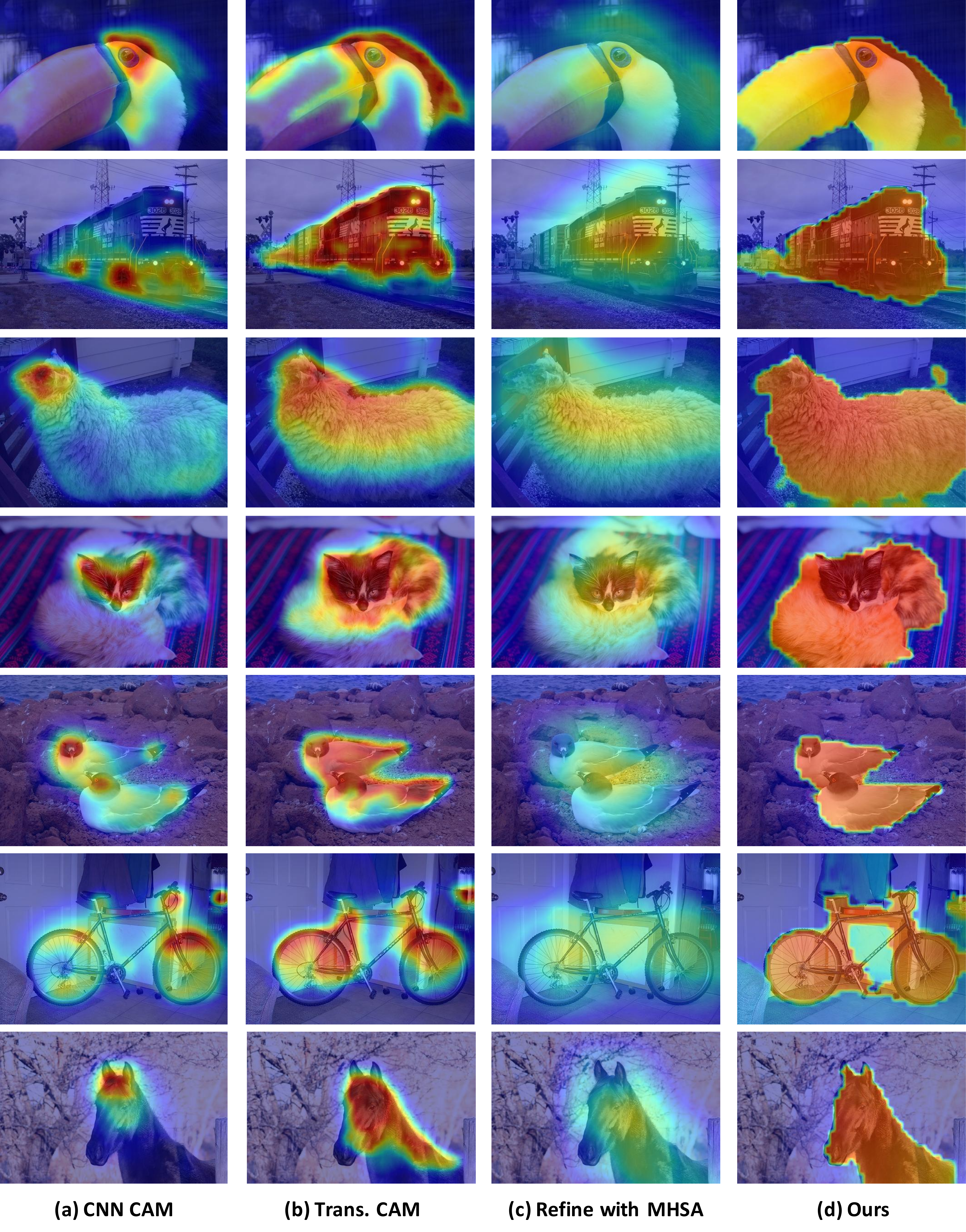}
    \caption{CAM generated with (a) Transformers activates more integral regions than (b) CNN. Refining CAM with (c) coarse MHSA doesn't work well, while (d) the learned affinity could remarkably improve the generated CAM.}
    \label{fig_intro_cam_sup}
\end{figure*}

\begin{figure*}[!tp]
    \centering
    \includegraphics[width=0.8\textwidth]{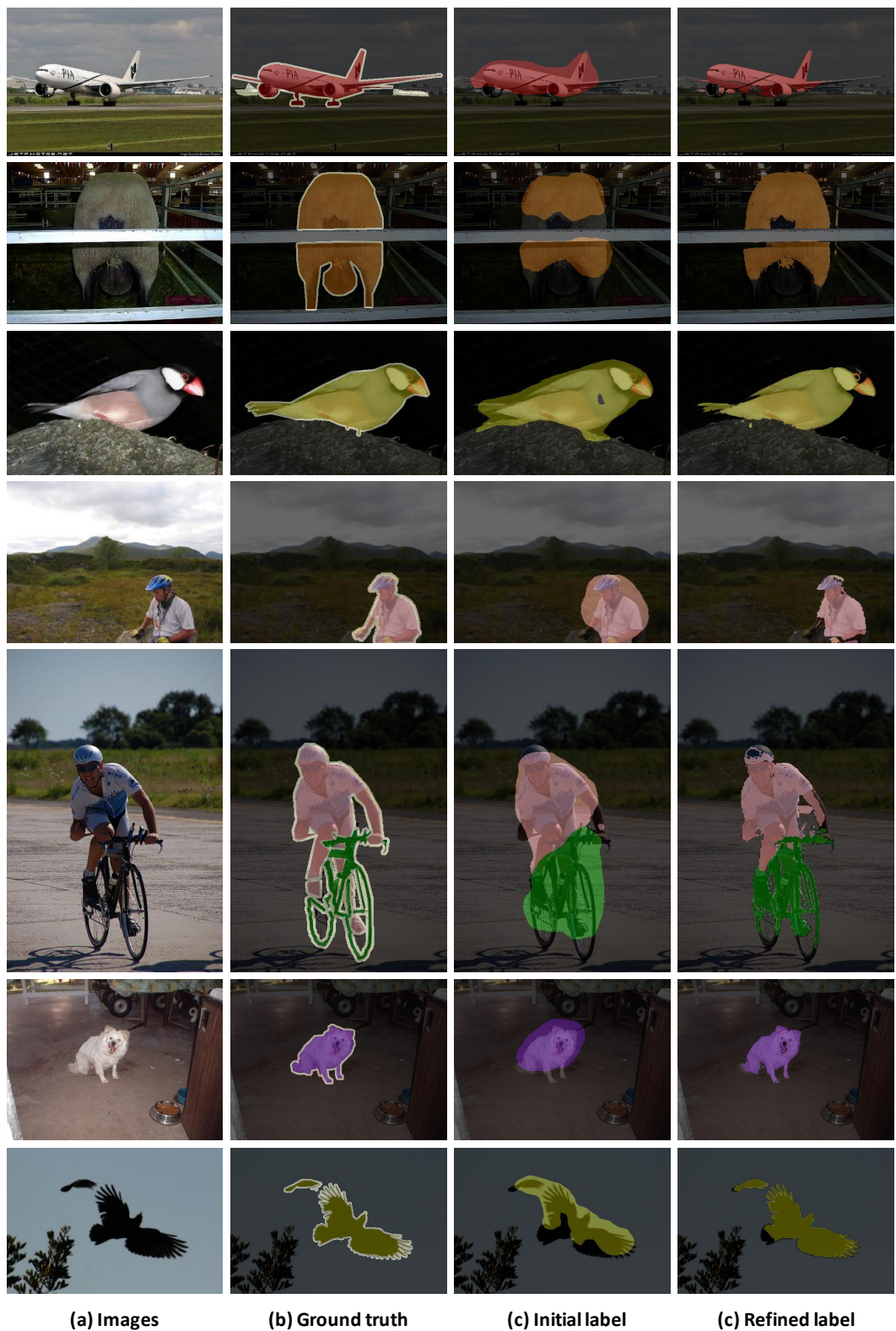}
    \caption{Improvements of the proposed pixel-adaptive refinement (PAR) module on the pseudo labels. The pseudo labels are generated with CAM and Transformer baseline. The proposed PAR could effectively dampen the falsely activated regions and ensure the alignment with low-level image appearance.}
\end{figure*}

\begin{figure*}[!tp]
    \centering
    \includegraphics[width=0.95\textwidth]{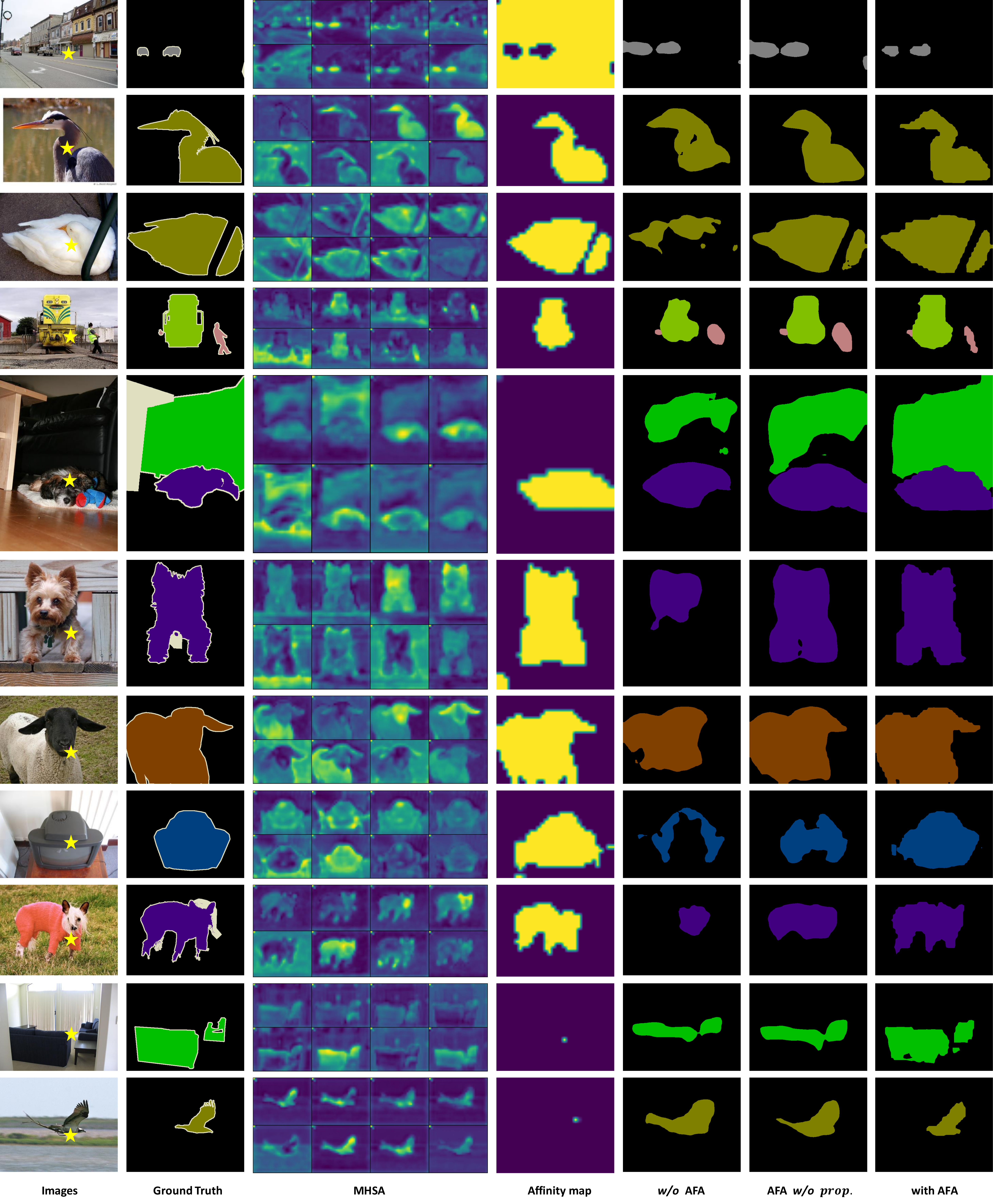}
    \caption{Visualization of the MHSA maps, learned affinity maps, and generated pseudo labels for segmentation. "$\bigstar$" denotes the query point to visualize the attention and affinity maps. The pseudo labels are generated with our model without AFA module (\textit{w/o} AFA), with AFA module but no random walk propagation (AFA \textit{w/o} $prop.$) and with full AFA module (with AFA). For the generated pseudo labels, the AFA module brings notable visual improvements. The affinity propagation process further diffuses the regions with high semantic affinity and dampens the regions with low affinity.}
\end{figure*}

\begin{figure*}[!tp]
    \centering
    \includegraphics[width=0.75\textwidth]{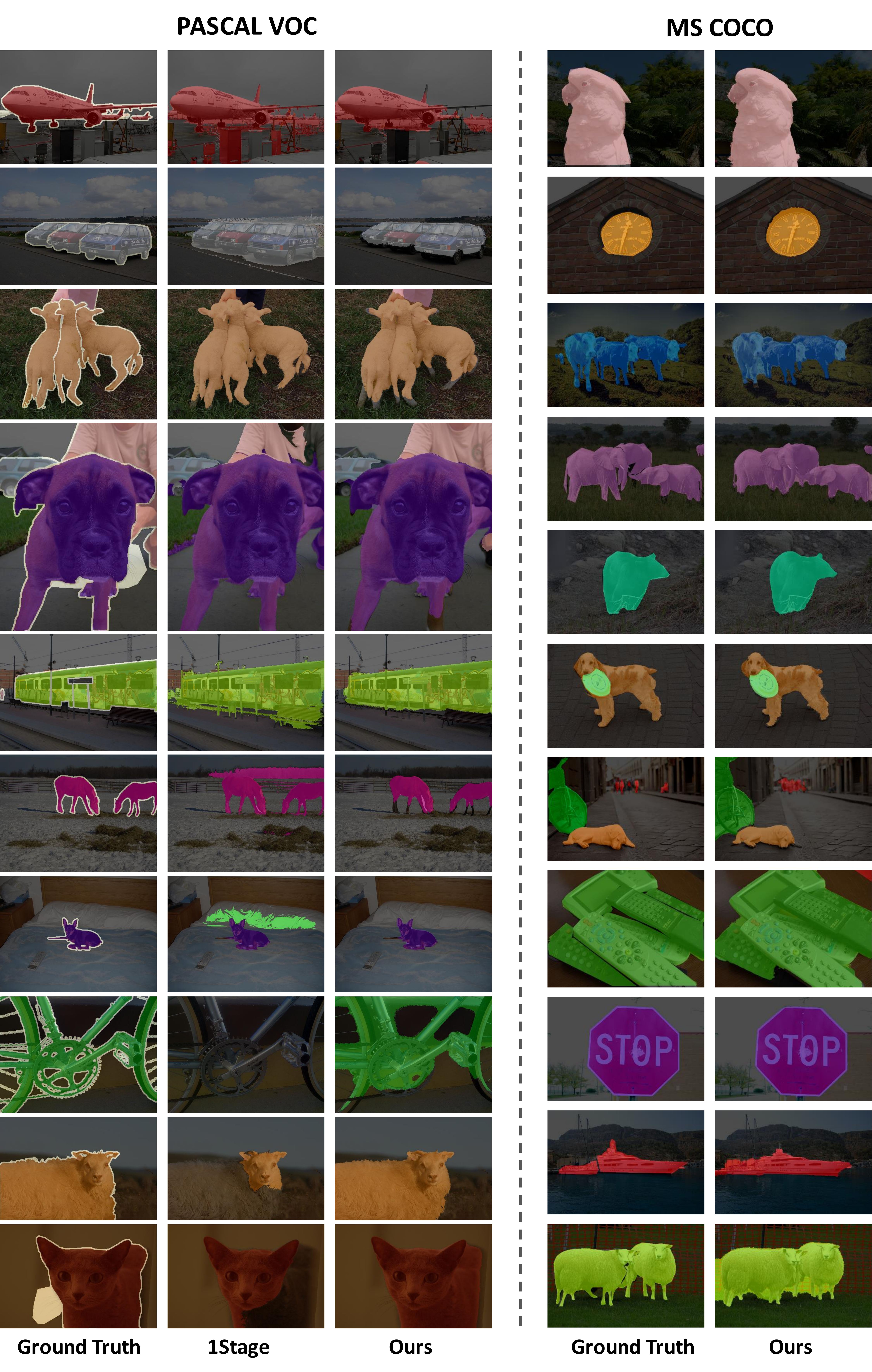}
    \caption{Semantic segmentation results on PASCAL VOC $val$ (left) and MS COCO $val$ set (right). Our method outperforms 1Stage \cite{araslanov2020single} and is comparable with ground truth labels.}
\end{figure*}

\begin{figure}[!tp]
    \centering
    \includegraphics[width=0.47\textwidth]{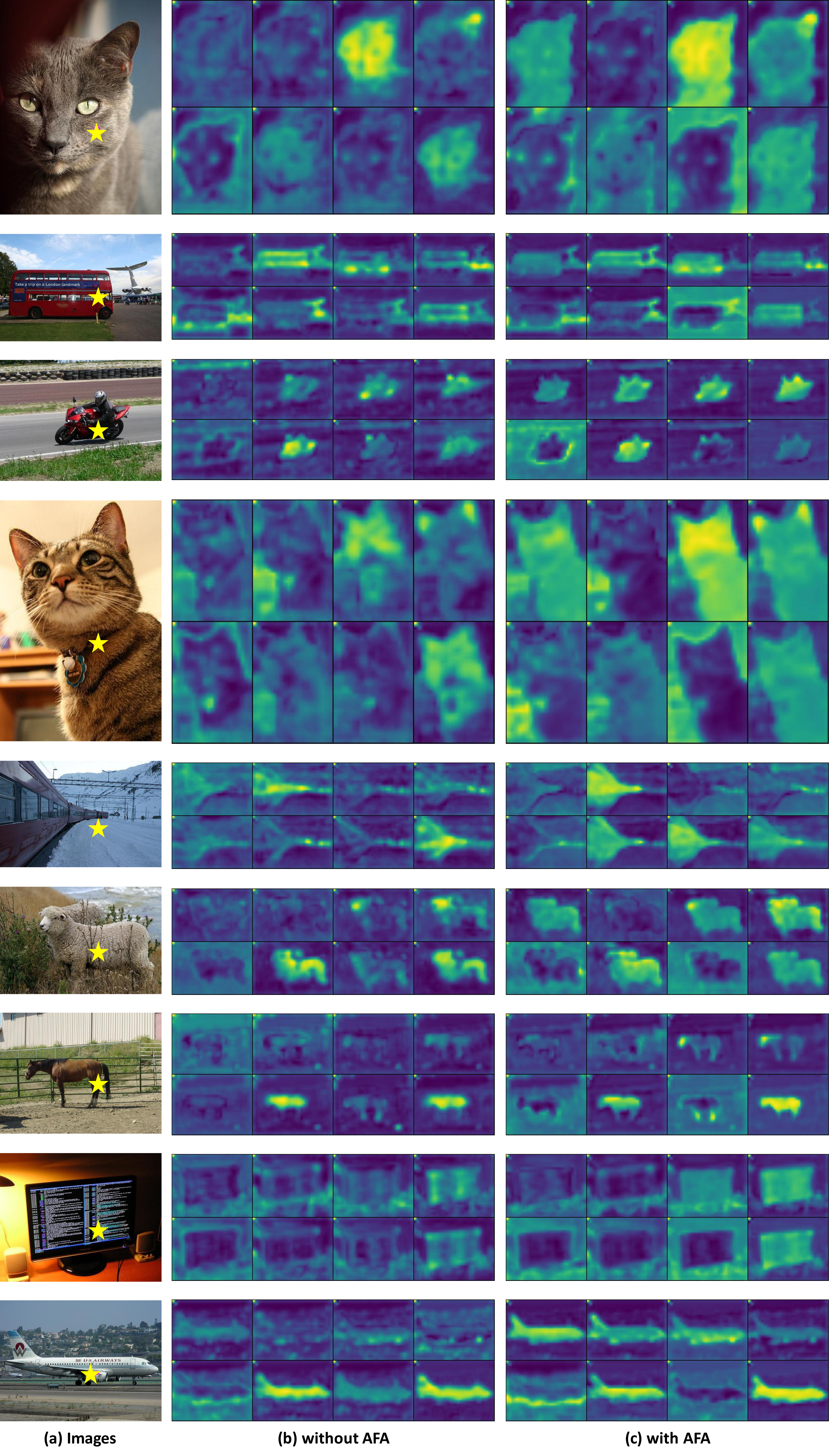}
    \caption{Visualization of the MHSA maps extracted from model without and with our AFA. "$\bigstar$" denotes the query point. Our AFA could help the MHSA to capture better semantic affinity.}
\end{figure}

\begin{figure}[!tp]
    \centering
    \includegraphics[width=0.47\textwidth]{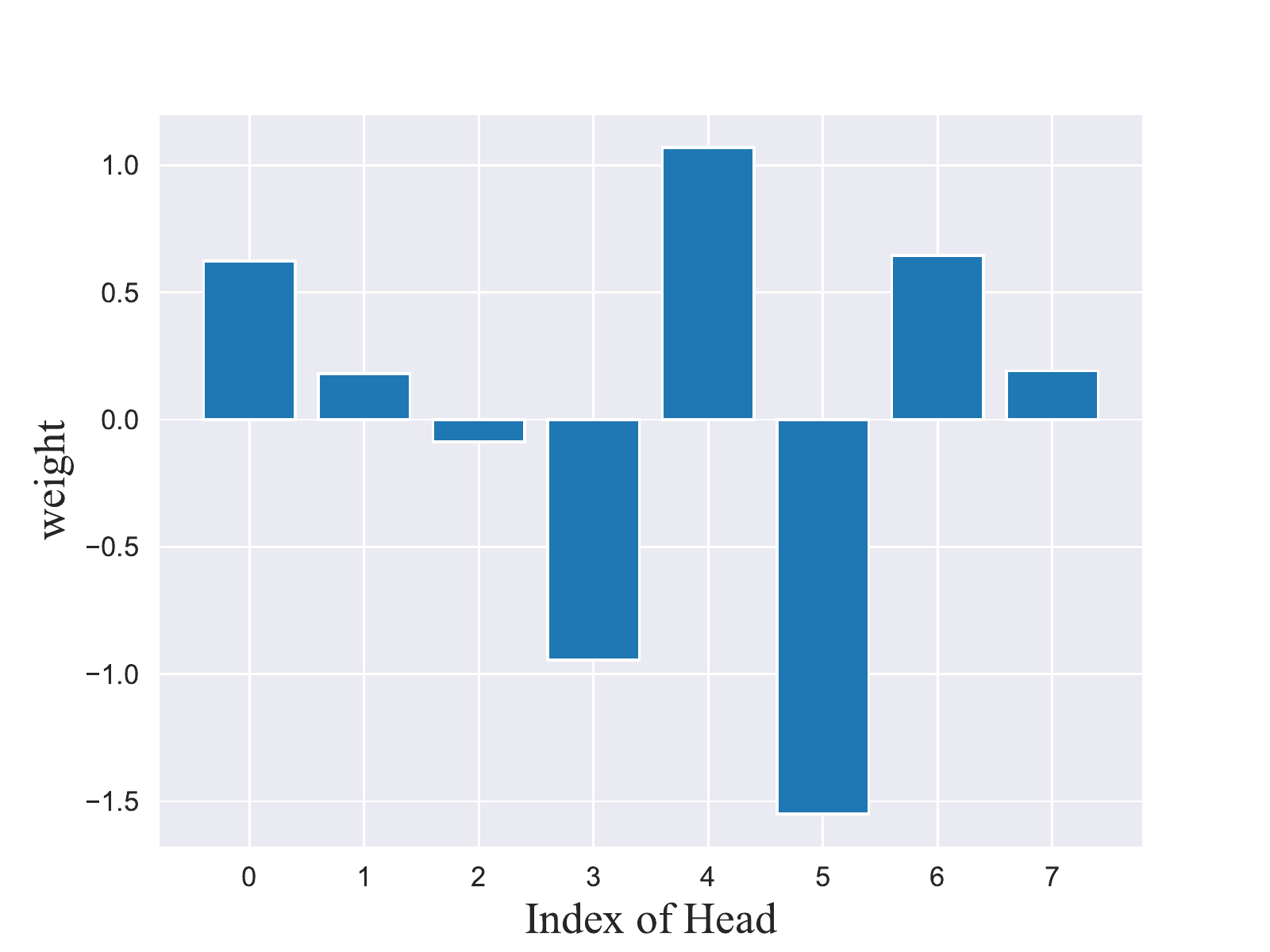}
    \caption{The learned weights of each head of self-attention in the AFA module. Here we only present the 8 heads of the last Transformer block. The MHSA matrices do not contribute equally to semantic affinity. Some self-attention matrices (head \#2, head \#3, and head \#5) contribute negatively to semantic affinity. The learned weights suggest applying MHSA directly as semantic affinity is not beneficial for the pseudo labels.}
\end{figure}

\end{document}